\documentclass[journal]{IEEEtran} 

\usepackage{amsmath}

\usepackage[dvipsnames]{xcolor}

\usepackage{float}

\usepackage{graphicx} 
\graphicspath{{images/}}
\DeclareGraphicsExtensions{.pdf,.jpeg,.png}

\usepackage[caption=false]{subfig}

\usepackage{multicol}
\usepackage{multirow}

\usepackage{url}

\usepackage{verbatim}

\begin{document}

\title{Open Evaluation Tool for Layout Analysis \\ of Document Images}

\author{Michele~Alberti,
        Manuel~Bouillon,
        Rolf~Ingold,
        Marcus~Liwicki \\ 
        University of Fribourg 
        Switzerland \\
        \{firstname\}.\{lastname\}@unifr.ch}

\markboth{}%
{Alberti \MakeLowercase{\textit{et al.}}: TODO INSERT TITLE HERE}

\maketitle

\begin{abstract}
%
This paper presents an open tool for standardizing the evaluation process of the layout analysis task of document images at pixel level.
We introduce a new evaluation tool that is both available as a standalone Java application and as a RESTful web service.
This evaluation tool is free and open-source in order to be a common tool that anyone can use and contribute to.
It aims at providing as many metrics as possible to investigate layout analysis predictions, and also provides an easy way of visualizing the results.
This tool evaluates document segmentation at pixel level, and supports multi-labeled pixel ground truth.
Finally, this tool has been successfully used for the ICDAR 2017 competition on Layout Analysis for Challenging Medieval Manuscripts.
\end{abstract}


\section{Introduction}
\label{toc:intro}


\setcounter{footnote}{1}

\begin{figure}[!t]
  \centering
  \subfloat[Original image.]{\includegraphics[width=\columnwidth,height=2.5cm]{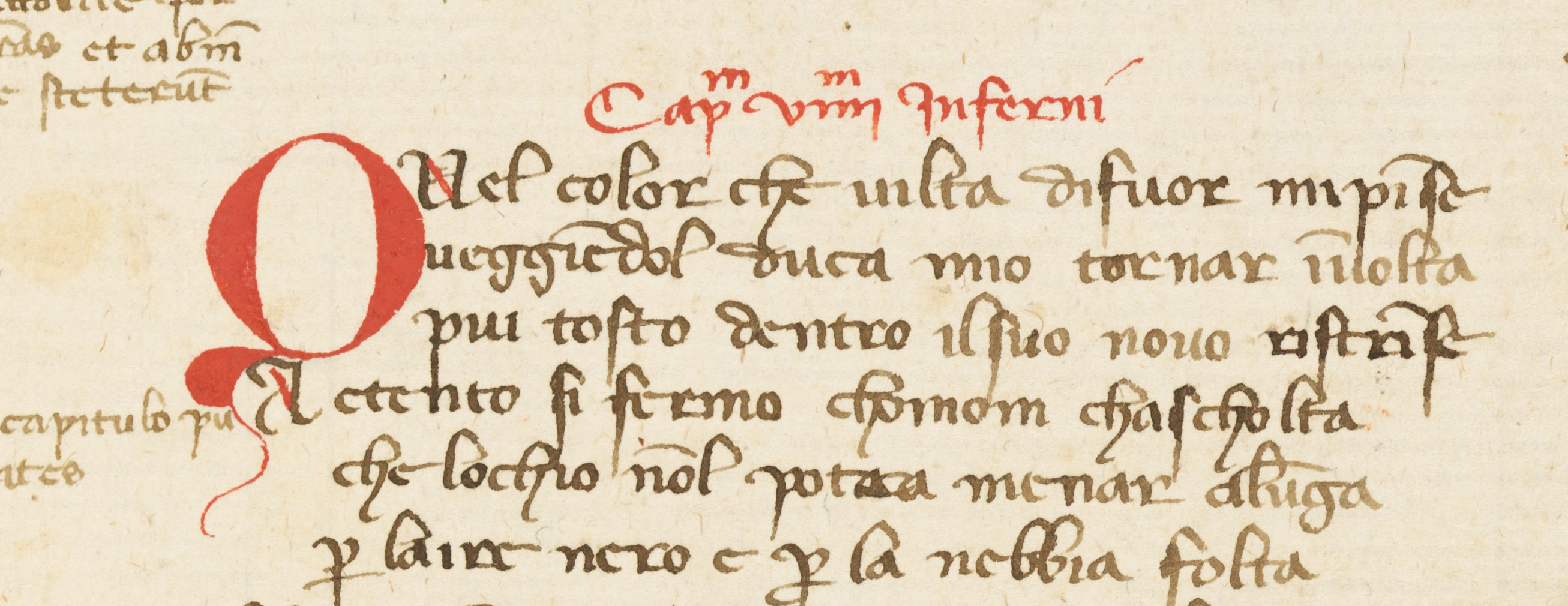}\label{subfig:original}}
  \\
  \subfloat[Prediction evaluation visualization]{\includegraphics[width=\columnwidth,height=2.5cm]{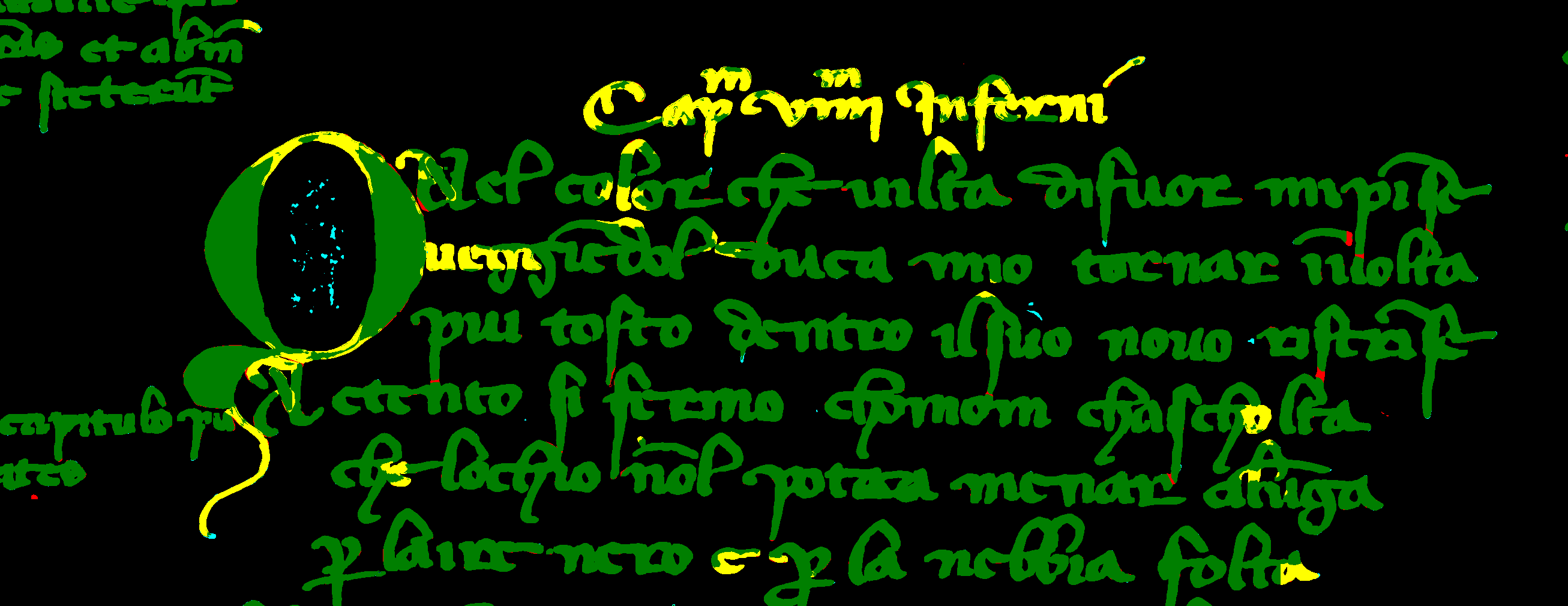}\label{subfig:visualization}}
  \\
  \subfloat[Overlap of prediction evaluation and original image]{\includegraphics[width=\columnwidth,height=2.5cm]{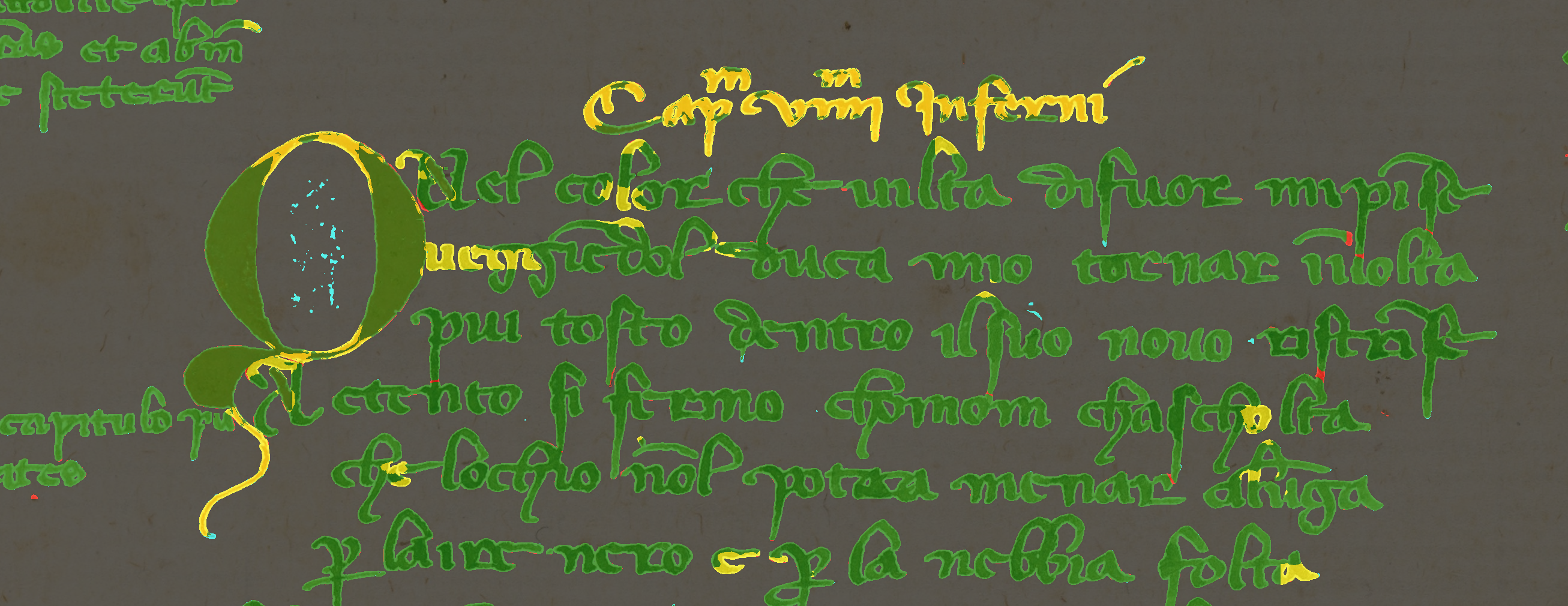}\label{subfig:overlap}}
  \caption[]{Exampleof output visualization produced by our tool for a cropped area of an image of CB55\footnotemark[1]. In the visualization (b) each pixel can be of five different colors: black for correctly classified background, red for background mis-classified as foreground, light-blue for foreground mis-classified as background, green for foreground pixels fully correctly classified and yellow for foreground pixels which are classified as such, but of the wrong class (e.g text instead of decoration).}
  \label{fig:visualization}
\end{figure}

\footnotetext{Cologny, Fondation Martin Bodmer, Cod. Bodmer 55: Dante, Inferno e Purgatorio (Codex Guarneri)  (\url{http://www.e-codices.unifr.ch/en/list/one/fmb/cb-0055}).}

Layout analysis, also known as document segmentation, is a common task in Document Image Analysis research.
Various evaluation schemes and metrics have been used throughout different competitions and benchmark reports.

The purpose of this work is to put in common the (typically additional) work needed to implement an evaluation scheme for layout analysis methods and have a reviewed source code to minimize the risk of erroneous implementations.
In a similar way that databases are shared publicly to enable the easy comparison of methods, we believe that evaluation tools should be publicly available as well.
It saves the necessity to implement something that already exists, and provide a well-tested implementation that is less likely to contains bugs.

In this paper, we present a new evaluation tool, that is free and open-source, for the evaluation of the layout analysis task of document images at pixel level.
It is available both as a standalone Java application and as a RESTful web service on DIVAServices~\cite{wursch_2015_divaservices} for easy integration in any programming language.
We strive making this tool exhaustive by implementing several state-of-the-art metrics.
Furthermore, we target at a reliable tool by having it well-tested and providing it open-source to enable peer reviews.
The tool also provides visualization capabilities to help understanding the results (see Figure~\ref{fig:visualization}), and the errors made by the evaluated method (see Figure~\ref{fig:visualization_error}).
This tool is also versatile in the sense that it designed to handle single-label as well as multi-label classification tasks.
Finally, this tool has been successfully used for the ICDAR 2017 competition on Layout Analysis for Challenging Medieval Manuscripts~\cite{simistira_2017_competition}.
Although the tool has been tested in the context of historical documents, it can be used in any other pixel-level classification scenario\footnote{This applies to both single and multi-label tasks.}, e.g. administrative documents and text detection in real scenes.

This paper is organized as follows.
Section~\ref{toc:related_works} discusses related work in evaluation metrics and evaluation tools for document image analysis.
Section~\ref{toc:tool} presents this new evaluation tool and how to use it.
Section~\ref{toc:metrics} describes the different metrics that are computed by this tool.
The visualization capabilities of this tool are also described in this section.
Section~\ref{toc:multi-label} explain the specificity of the multi-labeled classification task, and how the different metrics can be computed in this case.
Finally, Section~\ref{toc:conclu} concludes and discusses future work.

\section{Related Work}
\label{toc:related_works}

In this section, we present and discuss several existing typical metrics.
Furthermore, we summarize several approaches to make document analysis evaluation tools available at the moment.

The evaluation of layout analysis has been discussed by various researchers in this field and different metrics have been proposed.
Often, a simple pixel-accuracy is reported \cite{baechler2011multi, chen2014robust, alberti2017whatYouExpectIsNOTWhatYouGet, deng2009imagenet} which we regard as obsolete, because it is too much biased towards the majority class (see Section~\ref{toc:accuracy}).
Other works take the size of overlapping bounding boxes or regions into account~\cite{antonacopoulos2011historical,antonacopoulos2013icdar} or measure the Optical Character Recognition (OCR) performance on the resulting segmented page.

For providing Open Source tools and services, Neudecker et al. \cite{Neudecker2011} present an infrastructure providing access to a various range of tools for the whole OCR workflow.
The provided tools can be tried online for free\footnote{\url{https://www.digitisation.eu/tools-resources/demonstrator-platform/}}.
Full access to the Web Services is only available to paying members.
In the tranScriptorium project \cite{Sanchez2013}, the project members provide RESTFul Web Services\footnote{\url{https://transkribus.eu/wiki/index.php/REST_Interface}} for their methods as well and some tools on Github. However, their back-end infrastructure is not open. 

Targeting at the evaluation, Lamiroy and Lopresti~\cite{Lamiroy2011} introduced the Document Analysis and Exploitation (DAE) platform and a recent update called DAE-NG~\cite{lamiroy2017}.
The DAE-NG is a more broader effort, targeting the general synchronization of Document Analysis evaluation methods while we focus on layout analysis and provide various metrics as output.

The tool presented in this paper is available online (see Section~\ref{toc:tool}) and are furthermore integrated into the DIVAServices~\cite{wursch_2015_divaservices}.
Similarly, our tool can be integrated into DAE-NG or Transcriptorium as well.



\section{Evaluation Tool}
\label{toc:tool}

This tool was first used for the ICDAR 2017 competition on Layout Analysis for Challenging Medieval Manuscripts~\cite{simistira_2017_competition}.
We choose to make it freely available\footnote{\url{https://github.com/DIVA-DIA/LayoutAnalysisEvaluator}} to the DIA community in order have a common evaluation tool that can be used with the common databases we use.
Its goal is to enable the easy comparison of different layout analysis methods on the same database and with the same metrics.
This tool was designed for multi-label classification problems, as the competition, but it can also handle standard classification problems.
The interest and specificity of seeing document image layout analysis as a multi-labeled classification problem will be explained in Section~\ref{toc:multi-label}.

This tool is free and open-source\footnote{Under LGPL v3 licence.}, which means that anyone can use it, modify it and redistribute it.
Using this open-source tool permit to save the time that would be needed to re-implement what already exists, and to greatly reduce the risk of errors/bugs.
Indeed, this implementation of this tool has been proof-read, and anyone can check the correctness of the code online.
If an error or a bug would be found, it would be corrected and the correction easily shared with every user.

This segmentation analysis tool is implemented in Java, which is by definition multi-platform.
A JAR archive is available for download, and running it without parameters give the expected inputs:
\begin{itemize}
\item ground truth image: where the labels are encoded as for the database used for the competition\footnote{\url{http://diuf.unifr.ch/main/hisdoc/icdar2017-hisdoc-layout-comp}} (DIVA-HisDB~\cite{simistira_2016_diva}).
\item prediction image: same encoding as for the ground truth.
\item \protect{[output file]}: output values will be written in CSV format to this file if specified.
\item \protect{[output directory]}: visualization images will be generated to this folder if specified.
\end{itemize}
The output consists of various metrics, in order to evaluate the prediction and to permit comparisons with a maximum of published results.
As the tool is open-source, anyone can add new metrics if needed.
These different metrics will be detailed in Section~\ref{toc:metrics}.

To facilitate the integration of this evaluation tool in frameworks using different programming languages, we also make this tool available as a RESTful web service on DIVAServices\footnote{\url{http://divaservices.unifr.ch/api/v2/}}~\cite{wursch_2015_divaservices}.
First, a POST request with all the prediction images included in the JSON content (base64 encoded) should be sent to the following address \url{http://divaservices.unifr.ch/api/v2/collections} to upload all the prediction images into a collection.
Then, one just need to send a POST request to \url{http://divaservices.unifr.ch/api/v2/evaluation/icdar2017hisdoclayoutcomplineevaluation/1}, including the ground truth and the hypothesis image collection names as JSON content.
Finally, when the results are computed, DIVAServices send back the results with all the metrics in JSON format.

\section{Metrics}
\label{toc:metrics}

In this section, we present the different metrics measured by our tool and briefly discuss their interest. %
\par %
For each document the tool takes as input a ground truth image $L = (\{c \in C\}_i | 1 \leq i < n)$ and its relative predictions\footnote{The tool is designed to support multi-label classification problems. In the case of single-label problems, the elements of $L$ and $L'$ would not be a set but a single value.}  $L'$, where $C$ is the set of unique classes present in the document and $n$ is the total number of pixels in the document. %
The output is a set of metrics $\Psi = \{$\textit{exact match}, \textit{Hamming score}, \textit{Precision}, \textit{Recall}, \textit{F1-Score}, \textit{Jaccard index}$\}$ which aims to provide a variety of scores which should exhaustively describe the performance of the used algorithm. %
In fact, it is recommended to avoid trying to reduce the performance of a classification algorithm into a single score~\cite{yang_1999}. %
To this end, for each metric $\psi \in \Psi \setminus\{$\textit{exact match}, \textit{Hamming score}$\}$ we measure the value $\psi_c$ separately for each class $c \in C$ and additionally perform both macro-averaging and micro-averaging for evaluating the performance average across classes. %
Macro-averaged performance scores (denoted by $M$ superscript) are computed by simply averaging the class-wise scores, giving equal weight to each class, regardless of their size, as shown in Eq.~\ref{macro_averaging}. %

\begin{equation}
\label{macro_averaging}    
\psi^M = \frac{1}{|C|} \cdot \sum_{c \in C} \psi_c 
\end{equation}

Micro-averaged scores (denoted by the $\mu$ superscript) are obtained by averaging the class-wise scores a weight proportional to their frequency as shown in Eq.~\ref{micro_averaging}. %

\begin{equation}
\label{micro_averaging}    
\psi^\mu = \sum_{c \in C}  \psi_c \cdot f_c 
\end{equation}

where $f_c$ is the class-wise frequency computed as the cardinality of the set of pixels in the ground truth whose set of labels contain the class $c$, over the total amount of labels\footnote{Using the total amount of pixel here would be wrong, as in multi-label problem the summation of all frequencies would be higher than $1$.} in the ground truth, as shown below:

\begin{equation}
\label{frequency}
f_c = \frac{|\{p | c \in L_p \}|}{\sum |L_i|} 
\end{equation}

The evaluation is performed pixel-wise and for each of the classes involved at a single pixel there are four possible cases: %
\\
\begin{itemize}
    \item True Positive (TP): both prediction and ground truth contain the class label.
    \item True Negative (TN): both prediction and ground truth lack the class label.
    \item False Positive (FP): the prediction contains the class label, whereas the ground truth does not.
    \item False Negative (FN): the prediction lacks the class label, but the ground truth contains it.
\end{itemize}

\subsection{Exact Match}
\label{toc:exact_match}

The exact match is the most strict metric and literally counts how many exact matches there are between the predicted values $L'$ and the ground truth labels $L$. %

\begin{equation}
\label{exact_match}
EM = \frac{|\{p | L'_p = L_p \}|}{N} 
\end{equation}

The behavior of this metric tends to the behavior of accuracy when the average label cardinality tends to $1$. %
In the extreme case of single-class problem (where the average label cardinality is $1$) the exact match is equivalent to accuracy. %
This relationship makes the exact match sensitive to the same drawbacks as accuracy (see Section~\ref{toc:accuracy}). %

\subsection{Hamming Score}
\label{toc:hamming}

The Hamming score is a relaxed version of the exact match which accounts for partial matches\footnote{In case of single-label problem they are completely equivalent}. %
Therefore, it indicates the fraction of correctly predicted labels. %
It is computed as $1-$ \textit{Hamming loss} ($\oplus$ denotes exclusive or).

\begin{equation}
\label{hamming}
H = 1 - \frac{1}{n \cdot |C|} \cdot \sum | L'_i \oplus L_i |
\end{equation}

\subsection{Precision}

The precision of the classifier is defined as the probability that a random pixel predicted to belong to a class, 
really belongs to that class. %
In other words, $P_c$ denotes how many of the pixels predicted of class $c$ were actually belonging to class $c$. %

\begin{equation}
\label{precision}
P_c = \frac{TP_c}{TP_c + FP_c}
\end{equation}

\subsection{Recall}

The recall of a classifier is defined as the probability that a random pixel belonging to a class, will also be predicted to belong to that class. %
In other words, $R_c$ denotes how many of the pixels of class $c$ were actually predicted to be belonging to class $c$. %

\begin{equation}
\label{recall}
R_c = \frac{TP_c}{TP_c + FN_c}
\end{equation}

\subsection{F1-Score}

Since neither precision nor recall are conveying enough information to be used as evaluation metric alone, researchers developed different ways to combine them. %
One popular solution is the $F_\beta$ function which merges them by assigning $\beta$ times more importance to recall than to precision \cite{vanrijsbergen_1979}. %
The $F_1$-score\footnote{Also known as S\o rensen-Dice coefficient.} \cite{sorensen_1948} \cite{dice_1945} is a special case where precision and recall are given the same importance. %

\begin{equation}
\label{f1score}
F_{1c} = \frac{2 \cdot TP_c}{2 \cdot TP_c + FP_c + FN_c} = 2 \cdot \frac{P_c \cdot R_c}{P_c + R_c}
\end{equation}

The $F_1$-score corresponds to the harmonic mean between precision and recall. %

\subsection{Jaccard Index}
\label{toc:jaccard}

The Jaccard index often referred as Intersection over Union (IU) is a statistic used for comparing the similarity and diversity of sets \cite{levandowsky_1971}. %

\begin{equation}
\label{jaccard}
IU_c = \frac{TP_c}{TP_c + FP_c + FN_c}
\end{equation}

Unlike accuracy (see Section~\ref{toc:accuracy}) this metric is invariant to the total number of samples $n$ but sensible to changes in $FP$ and $FN$. %
Additionally, as one can see, it is computed in a similar way to the $F_1$ score (see Equations~\ref{f1score} and~\ref{jaccard}). %
The Jaccard index is, however, much more sensitive to mistakes as it is not weighting twice the true positive cases. %
This makes it effectively the strictest metric in our set $\Psi$, and despite we advocate against using a single metric to compare two algorithms performances, we recommend to use the Jaccard index if one were to do it anyway. %

\subsection{About accuracy}
\label{toc:accuracy}

Accuracy is a common performance measure in the machine learning literature, but there is a potential pitfall if used to evaluate the performances in a multi-class setting~\cite{yang_1999,sebastiani_2002}. %
In fact, as there is $n$ in the denominator (see Eq.~\ref{accuracy}), a small change in $TP_c$ or $TN_c$ will practically not affect the final result if $n$ is large enough. %
Additionally, if there is a high number of classes or if class sizes are small, the trivial rejector\footnote{The trivial rejector is a classifier that systematically rejects all classes.} might have a surprisingly high accuracy (close to $100\%$) because of $TN_c \approx n$. %
Since in the context of document segmentation at pixel level these conditions are met --- high resolutions documents have a large $n$ and some classes only have a few representatives --- we consider accuracy not as interesting as the other metrics aforementioned and we consequently omit\footnote{Note that in case of single-class problem, the accuracy is equivalent to both the exact match and the Hamming score (see Sections~\ref{toc:exact_match} and~\ref{toc:hamming}).} it in our tool. %

\begin{equation}
\label{accuracy}
ACC_c =  \frac{TP_c + TN_c}{n}
\end{equation}

\subsection{Visualization}

\begin{figure}[!t]
  \centering
  \subfloat[Prediction quality visualization ]{\includegraphics[width=.45\columnwidth,height=4cm]{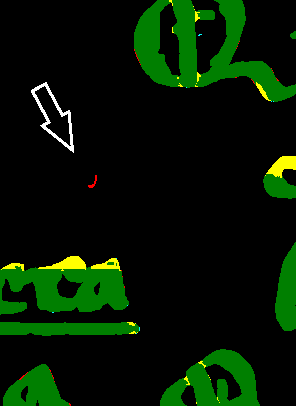}\label{subfig:visualization_error}}
  \hfil
  \subfloat[Overlap of prediction quality and original image]{\includegraphics[width=.45\columnwidth,height=4cm]{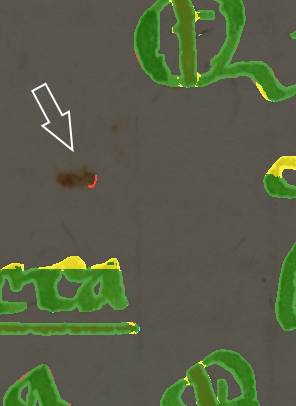}\label{subfig:overlap_error}}
  \caption{Example supporting the usefulness of overlapping the prediction quality visualization with the original image. Focus on the red pixels pointed at by the white arrow: they are background pixels misclassified as foreground. In the normal visualization (a) it's not possible to know why would an algorithm decide that in that spot there is something belonging to the foreground, as it is clearly far from regular text. However, when overlapped with the original image (b) one can clearly see that in this area there is an ink stain which could explain why the classification algorithm is deceived into thinking these pixels were foreground. This kind of interpretation is obviously not possible without the information provided by the original image like in (a).}
  \label{fig:visualization_error}
\end{figure}

Once a document is evaluated, the user receives a relatively huge\footnote{The exact number is $2 + (|\Psi|-2) \cdot (|C|+2) =  4\cdot |C| + 10$, which is obtained by concatenating \textit{exact match}, \textit{Hamming score} and $|C|$ class-wise plus micro and macro averaging for all $|\Psi| -2$ remaining metrics.} amount of numbers as output, and we felt that might be not only overwhelming but also difficult to interpret at times. %
Therefore we included a visualization method which allows the user to look at the quality of the prediction through colors and not through numbers (see Figure~\ref{subfig:visualization}). %
If desired, this visualization can be overlapped to the original image to ultimately empower the user to estimate the results with his own eyes (see Figure~\ref{subfig:overlap}). %
This kind of representation enables the interpretation of the segmentation mistakes by displaying exactly which pixels have been misclassified, which is clearly not possible with any numerical metric (see Figure~\ref{fig:visualization_error}). %

\section{Multi-Label}
\label{toc:multi-label}

\begin{figure}[t]
  \centering
  \includegraphics[width=.45\columnwidth]{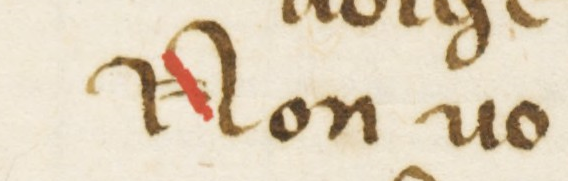}
  \caption{Example of multi-label dataset, where the red pixels forming the ``N'' are clearly both part of a decoration and part of the text. This image is extracted from the same dataset used in Figure~\ref{fig:visualization}.}
  \label{fig:multi-label}
\end{figure}

This tool has been designed to support not only multi-class but multi-label classification problems as well. %
This is a mandatory requirement to properly evaluate the datasets which have a multi-label ground truth (e.g. the DIVA-HisDB dataset~\cite{simistira_2016_diva}). %
This kind of dataset is interesting in the context of historical documents as a foreground pixel might belong to several classes at the same time. %
For example, decorations are frequently used as part of the text, i.e drop caps. %
In that case, the pixels would belong to both text and decoration. %
Another example is shown in Figure~\ref{fig:multi-label} where the red pixels forming the ``N'' are both part of a decoration and part of the text. %

The problem that rises going from single to multi-label classification is that computing the confusion matrix is no longer straightforward. %
Although it is technically possible to build one\footnote{One can extend the regular confusion matrix with a row/column for each possible combination of the classes in order to cover all possible cases.}, there are two major drawbacks. %
First, the size of such matrix would be $2^{2n}$ which is impractical (arguably impossible) for a human to efficiently interpret. %
Second, extracting $TP$, $TN$, $FP$ and $FN$ from such a matrix is much more complicated both from the theoretical and practical point of view than in the regular single-label confusion matrix. %
To overcome this issue, we compute the contingency table for all classes separately, like in the example below: %

\begin{table}[H]
\centering
\begin{tabular}{lccc}
                            &                                         & \multicolumn{2}{c}{\multirow{2}{*}{Ground Truth}}      \\
                            &                                         &                         &                              \\
                            & \multicolumn{1}{c|}{}                   & \multicolumn{1}{c|}{A}  & $\overline{A}$               \\ \cline{2-4} 
                            & \multicolumn{1}{c|}{\multirow{2}{*}{A}} & \multicolumn{1}{c|}{\multirow{2}{*}{TP}}   & \multicolumn{1}{c}{\multirow{2}{*}{FP}} \\
\multirow{2}{*}{Prediction} & \multicolumn{1}{c|}{}                   & \multicolumn{1}{c|}{}   &  \\ \cline{2-4} 
                            & \multicolumn{1}{c|}{\multirow{2}{*}{$\overline{A}$}} & \multicolumn{1}{c|}{\multirow{2}{*}{FN}} & \multicolumn{1}{c}{\multirow{2}{*}{TN}} \\
                            & \multicolumn{1}{c|}{}                   & \multicolumn{1}{c|}{}   &  \\  
           
\end{tabular}
\label{tab:cm}
\end{table}

This way we have $|C|$ different matrices of constant size $4$ to which is straightforward to apply the formulas presented in Section~\ref{toc:metrics}. %


\section{Conclusion}
\label{toc:conclu}

This paper promotes a tool that is very easy to use for evaluating document image segmentation at pixel level.
It is open-source and freely available, both as a standalone application and as a RESTful web service.
This tool is designed to correctly support single and multi-label problems, as well as single-class ones.
The numerical output metrics are numerous and reliable, and we strongly suggest the use of the Jaccard Index (Intersection over Union) for a first global evaluation.
Additionally, we generate a very human-friendly visualization of the results which allows performing jointly a quick estimation of the prediction quality and a deep inspection of its mistakes.
We plan to extend this tool in the near future to allow users to provide polygons as input data in the PAGE XML format.  


\section*{Acknowledgment}
The work presented in this paper has been partially supported by the HisDoc III project funded by the Swiss National Science Foundation with the grant number $205120$\textunderscore$169618$.

\bibliographystyle{IEEEtran}
\bibliography{biblio}

\end{document}